\documentclass[a4 paper, 11pt,twoside]{article}
\usepackage{fullpage}                           
\usepackage[lmargin=0.6in,rmargin=0.6in,tmargin=0.6in,headsep=.2in]{geometry}
\usepackage{graphicx}
\usepackage{forloop}
\usepackage{etoolbox}
\usepackage{calc}
\usepackage{wrapfig,lipsum,booktabs} 
\usepackage{xtab} 
\usepackage[dvipsnames]{xcolor}
\usepackage[normalem]{ulem}
\usepackage{sectsty}
\allsectionsfont{\color{Brown}}
\makeatletter
\def\@seccntformat#1{\@ifundefined{#1@cntformat}%
{\csname the#1\endcsname\;}
{\csname #1@cntformat\endcsname}
}
\def\section@cntformat{\thesection.\;} 
\def\subsection@cntformat{\thesubsection.\;} 
\makeatother
\usepackage{hyperref}
\hypersetup{
    colorlinks=true,
    urlcolor=blue,
    linkcolor=black,
    citecolor = black}

\usepackage{fancyhdr}
\fancypagestyle{first}{%
  \fancyhf{}
  \fancyfoot[L]{{\footnotesize \textsf{DOI: \href{10.4236/jcc.2020.89001.2020.89001}{\color{blue}\uline{10.4236/jcc.2020.89001}} $\;$Sep. 04, 2020}}}%
  \fancyhead[R]{{\bf\small \textsf{Journal of Computer and Communications, 2020, 08, 1-16}}\\
\href{http://www.scirp.org/journal/jcc}{\color{blue}\uline{\textsf{http://www.scirp.org/journal/jcc}}}\\
\textsf{ISSN Online:2327-5227}\\
\textsf{ISSN Print:2327-5219}}%
}
\usepackage{amsmath,amsfonts,amssymb,amsthm,epsfig,epstopdf,url,array}
 \usepackage[retainorgcmds]{IEEEtrantools}
 \usepackage{makeidx,epsfig,lscape}
 \usepackage{xcolor,pict2e}
\usepackage{upgreek}

\theoremstyle{definition}

\begin{document}
\thispagestyle{first}
\vspace*{3cm}
{\noindent\huge\bf A Methodological Approach to Model CBR-based Systems}\\[1cm]
{\bf\large Eliseu M. Oliveira\textsuperscript{1}, Rafael F. Reale\textsuperscript{1}\textsuperscript{2} and Joberto S. B. Martins\textsuperscript{1}}\\[0.5cm]
\textsuperscript{1}\small Universidade Salvador - UNIFACS, Salvador, Brazil\\
\textsuperscript{2}Instituto Federal da Bahia - IFBA, Valença, Brazil\\
Email: eliseu@gmail.com, reale@ifba.edu.br, joberto.martins@unifacs.br\\
\begin{wraptable}{l}{5.1cm}
{\footnotesize
\begin{xtabular*}{0.3\textwidth}{p{5cm}}
\noindent{\bf How to cite this paper:} Oliveira, M. E.; Reale, R. F.; Martins, J. S. B. (2020) A Methodological Approach to Model CBR-based Systems, Journal of Computer and Communications, {08},    1-16.\\
\url{https://dx.doi.org/10.4236/jcc.2020.89001 .2020.89001}\\
\\
{\bf \tiny Received: July 22, 2020}\\
{\bf \tiny Accepted: September 01, 2020}\\
{\bf \tiny Published: September 04, 2020}\\
\\
Copyright \copyright$\;$2020 by author(s) and Scientific Research Publishing Inc.\\
This work is licensed under the Creative Commons Attribution International License (CC BY 4.0).\\
\url{http://creativecommons.org/licenses/by/4.0/}\\
\includegraphics[width=2.5cm,height=0.72cm]{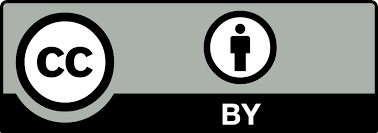}$\;$\includegraphics[width=2.5cm,height=0.75cm]{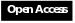}\\
{\color{white}\lipsum[1-60]}
\end{xtabular*}
}
\end{wraptable}
{\color{Brown}\rule{0.7\textwidth}{2pt}}\\[0.2cm]
{\color{Brown}\bf\large Abstract}\\
\\
Artificial intelligence (AI) has been used in various areas to support system optimization and find solutions where the complexity makes it challenging to use algorithmic and heuristics. Case-based Reasoning (CBR) is an AI technique intensively exploited in domains like management, medicine, design, construction, retail and smart grid. CBR is a technique for problem-solving and captures new knowledge by using past experiences. One of the main CBR deployment challenges is the target system modeling process. This paper presents a straightforward methodological approach to model CBR-based applications using the concepts of abstract and concrete models. Splitting the modeling process with two models facilitates the allocation of expertise between the application domain and the CBR technology. The methodological approach intends to facilitate the CBR modeling process and to foster CBR use in various areas outside computer science.
\vspace{2cm}\\
{\color{Brown}\bf\large Keywords}\\
\\
Artificial Intelligence; Case-based Reasoning; CBR Modeling; Bandwidth Allocation Model.
\vspace{0cm}\\
{\color{Brown}\rule{0.7\textwidth}{2pt}}

\normalsize
\section{Introduction}{}
\label{sec:BAM-CBR}

Artificial Intelligence (AI) and Machine Learning (ML) techniques are being extensively used in an ever-increasing number of areas and systems. They provide benefits on adopting them, such as efficient optimization methods and the possibility to solve rather complex multi-objective and multi-constrained problems that were difficult or eventually impossible to solve with current algorithmic or heuristics solutions \cite{boutaba_comprehensive_2018}.

ML-assisted applications are a trend, and many researchers and developers are rushing to apply ML and recover their inherent potential benefits \cite{adadi_peeking_2018} \cite{mahdavinejad_machine_2018}.

However, using ML techniques to solve any problem do require some previous background and expertise. For example, it is vital to choose the ML technique that better suits the target application in terms of available computational capability and expected target results.  In sequence to an adequate ML technique choice, it is typically necessary to model the problem under the premises of the chosen technique. The modeling process may include, as an example, an  MDP-based markovian process (Markov Decision Process) like Q-Learning or SARSA  formulation for Reinforcement Learning  or the definition of a neural network structure for Neural Networks (NN) \cite{sutton_reinforcement_1998} \cite{parisi_continual_2019}.

Case-based Reasoning (CBR) \cite{Aamodt1994} is a technique for problem-solving and for capturing new knowledge (learning) based on the stored knowledge of past experiences. CBR paradigm has a base of past experiences, called a case-base, and attempts to solve new problems by recovering similar solutions in this database and adapting them to new problems. CBR, to some extent, mimics the human behavior in activities like management and diagnostics in which the previous knowledge and experience is the driver in looking for the solutions for new near-equivalent problems \cite{perner_case-based_2014} \cite{boukehila_case-based_2019}.

CBR was proposed more than a decade ago as an AI technique \cite{Aamodt1994}, is simple to use, has minimal learning requirements, and does not typically require intensive computational resources \cite{khan_hybrid_2019}. More recently, CBR receives the attention from the artificial intelligence community and is gaining track in domain areas like medicine, expert systems, retail, smart grid, construction, manufacturing, design, agriculture and management  \cite{weglarz_application_2018} \cite{osuszek_case_2015} \cite{su_developing_2019} \cite{choudhury_survey_2016} \cite{lopez-fernandez_using_2011} \cite{calhau_electric_2019}.


Like other AI techniques, CBR requires the target system to be modeled to allow a similarity search in its database. This process is not clearly detailed or methodologically described in the literature. Our approach provides a way to methodologically model the target system. In our approach, the CBR modeling for problem-solving requires a specialized abstract model of the target system and, derived from it, a concrete CBR representation of the variables and parameters involved in the process.

The objective of this work is, in summary, to propose a methodological approach to model the CBR process based on a mapping between the abstract and concrete representations of the CBR process variables and parameters. The proposed method aims to facilitate the CBR modeling process and contribute to promoting the widespread use of CBR. We also expect the contribution can be relevant to CBR application areas where the computer science expertise of the professionals involved is less frequent or even unavailable.

The remaining of this paper is organized as follows. Section \ref{sec:fundamentals} presents an overview of the CBR fundamentals and section \ref{sec:work}  discusses the related works. Sections \ref{sec:RepresentacaoAbstrataCBR}  and \ref{sec:RepresentacaoConcretaCBR} present the CBR modeling methodology and section \ref{sec:CasoUso} follows with an example of how to use the approach for the cognitive management of bandwidth in network links. In section \ref{sec:conclusion}, the final considerations are presented.

\section{Case-based Reasoning Fundamental Aspects}\label{sec:fundamentals}

In summary, in CBR, a new problem is solved through the knowledge and information available or acquired by a previous similar problem adapted to create a new solution.

The essential components used to solve a problem with CBR are a \textit{case}, the \textit{case-database} of past cases or past experiences, and the \textit{similarity function}.

The \textit{case} is the way we represent the experience we have about the target problem. All previous \textit{cases} representing the acquired knowledge are stored in a database, called \textit{case-database}. A \textit{case} is represented by a pair \textit{problem and solution} that are the fundamental aspects present in all CBR systems:

\begin{itemize}
    \item \textit{Problem}: It contains all the information regarding the past event that you want to remember. That is, it describes all the essential data for the representation of knowledge in the specific domain. These data can include contextualization data, application objectives, descriptions of what happened, qualitative data, and quantitative data, among others.
    \item \textit{Solution}: It presents the necessary information to solve the problem related to the past event. This solution can be any information or action that totally or partially solves the problem presented in the case description. The representation of the solution must always take into account the application domain.
\end{itemize}

\begin{figure}[ht]
\centering
\includegraphics[width=.5\linewidth]{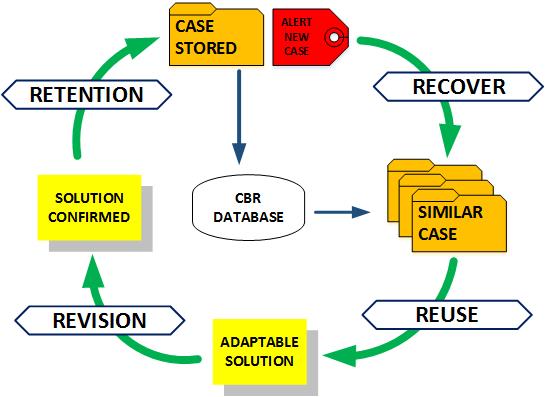}
\caption{CBR 4R Cycle \cite{m._oliveira_evaluating_2017}}
\label{fig:4R}
\end{figure}

The execution of the CBR process includes 4 phases using its essential components, namely  (Figure \ref{fig:4R}) \cite{m._oliveira_evaluating_2017} \cite{Wangenheim_2013}:

\begin{itemize}
    \item \textit{Recover}: Performs the search for similar cases in the case-database. This comparison is made using the similarity function. The similarity function is responsible for comparing the base cases with the actual problem and should return the most similar cases found.
    \item \textit{Reuse}: The phase in which the description of the current problem is composed with the solution of the case recovered in the previous phase. Then the solution found is applied to the environment in question.
    \item \textit{Revision}: In the review phase, the specialist in the technological domain must assess whether the solution employed brings the expected results. The specialist has the opportunity to make fine adjustments to optimize, adjust, or adapt the recommended solution.
    \item \textit{Retention}: After making the necessary adaptations, the specialist must confirm the new case as a valid case and consequently save it in the case base for later reuse.
\end{itemize}

From the methodological point of view, the utilization CBR to solve a particular problem involves the following modeling and operational steps:

\begin{itemize}
    \item \textit{Case and Knowledge Representation}: The case components and its structure embed the experience and the knowledge for a particular problem. As such, case descriptions document real experiences, and the CBR system may acquire new knowledge by retaining new cases.

    \item \textit{Similarity Measure}: A similarity function calculates the similarity measure, and it defines to what extent a case in the case-database is similar to the case being processed.

    \item \textit{Adaptation}: In CBR, the operational adaptation step takes a similar case and adapts it to the current situation. In general, adaptation uses a defined mechanism, expert knowledge, or a mix of both.
    \item \textit{Learning}: It is an operational step that allows the CBR system to memorize its successful and unsuccessful solutions, which means effectively to acquire new knowledge.

\end{itemize}

The foundation of the CBR that supports its capability to adapt, learn, and retain knowledge are the case representation and the choice of similarity measures. Accurately modeling these two elements of the CBR operation is essential.

In this paper, we present a methodological approach that addresses the representation of cases and the choice of similarity measures for the problem of resource management with CBR.

The proposed methodological approach is composed of an Abstract Model (AM) and a Concrete Model (CM). The abstract model is a high-level representation that structures the knowledge corresponding to the scope of actuation. The concrete model maps the abstract model's representation to the set of parameters used in the CBR execution process to acquire knowledge for a set of particular cases.

\section{Related Work}\label{sec:work}

Watson presents a discussion about the methodological approach used by CBR in \cite{watson_case-based_1999}. Watson argues that CBR can be better described as a methodology for problem-solving and, as such, differs from other artificial intelligence techniques like neural networks, and genetic algorithms that use more formal mathematical methods. Watson's paper does not discuss or propose a methodological approach to support CBR deployments.

The case representation formalism is discussed in Martines \cite{zhai_associated_2019}. The authors  discuss how experiences can be represented by using since simple feature vectors to representational formalisms like object-oriented, predicate-based and semantic nets, among others.

Althof discusses in \cite{bergmann_methodology_1998} the need to develop CBR applications as a systematic engineering activity. The paper addresses the software development life-cycle targeting the development of software products that support CBR development.

A discussion about CBR modeling is presented in Krite \cite{weber_cbr_2005}. The paper describes how CBR can be used to compare, reuse, and adapt inductive models that represent complex systems. The paper does not address specific cases and knowledge representations.

\section{Abstract Model}
\label{sec:RepresentacaoAbstrataCBR}

The proposed approach to model a CBR-based system for problem-solving has two steps:

\begin{itemize}
    \item Abstract Model (AM) definition; and
    \item Concrete Model (CM) mapping.
\end{itemize}

The definition of the AM requires expertise in the application domain involved. The mapping to the CM from the AM requires a minimum CBR expertise.

The abstract model is a high-level representation to structure the knowledge and, as such, to define the scope of action of the problem-solving system. It requires specialist knowledge of the application domain to which a system using CBR is applied. This model aims to represent knowledge about the scope of actions for the CBR system.

The abstract model is composed by a Technological Domain (TD) with general and specific objectives, attributes, measurements, actions, and premises.

\subsection{Technological Domain (TD)}

The technological domain defines, in general, the scope of the target problem-solving or, in other words, what the problem is and how to represent it \cite{m._oliveira_evaluating_2017} \cite{Wangenheim_2013}.

The representation of the TD in the proposed model is as follows:

\begin{itemize}
    \item System and problem description and objectives;
    \item Static, contextual and dynamic attributes of the system;
    \item Measurements; and
    \item Actions.
\end{itemize}

This set of components adequately describes the target system for general resource management problems like virtual machine management, datacenter management, and network management to cite some application examples.

\subsection{TD Component - System and Problem Definition,  General and Specific Objectives}

The \textit{system and problem description} is a formal or textual description of CBR problem and system.

The \textit{general objective} is a high-level definition and delimitation of the target management problem. It is associated to the CBR resource management task.

The \textit{specific objectives} detail the \textit{general objective} of the CBR management system's multiple possible outcomes. In effect, this corresponds to the definition of a multi-objective problem, possibly under a multi-requirement scenario.

The \textit{specific objectives} specify a subset of the context, measurement, and action attributes used in the CBR decision-making and learning processes.

\subsection{TD Component - Attributes}

The \textit{static attributes} describe static characteristics of the target system and have, in most cases, documentary value. The CBR system operation does not index the static attributes, so they do not interfere in the problem-solving process.

The \textit{contextual attributes} indicate the context of the problem and correspond to definitions and parameters that do not change frequently. From the CBR system perspective, contextual attribute modification may imply in restarting the learning process with CBR.

The \textit{dynamic attributes} are the set of variables and parameters that indicate the target system's global state. In the problem-solving context, dynamic attributes impact the target management process and are measurable.

\subsection{TD Component - Measurements, Tolerance and Actions}

The \textit{measurements} are, in general, the set of variables actual values acquired by a monitoring system that are relevant to the management process. The measurements, together with the context attributes, provide a snapshot of the system current state.

The measurements are instances of the system variables and aims to have a snapshot of the systems and quantify the specific objectives. In effect, the management of accepted values for measurable variables quantify, in general, the specific objectives.

\textit{Tolerance} represents the accepted value for the scope of this methodology. So, in most cases, the specific objectives are represented by ranges, upper or lower limits for defined managed variables.

The \textit{actions} are the operation set that is used to react upon the identification of a problem. The set of actions defined is executed on the system and are related to the defined specific objectives.

\subsection{Premisses}

In the CBR abstract model, the premises are the set of problems to which a solution is known. Premises are optional and aim to facilitate knowledge acquisition by the CBR system. It is essential to highlight that the CBR system is capable of learning from scratch without any given premise.

The premises can be confirmed or not, through the acquisition of new knowledge. Wrong premises will negatively impact the learning process since the premise will be discarded to allow the acquisition of new knowledge.

\section{Concrete Model}
\label{sec:RepresentacaoConcretaCBR}

The concrete model corresponds to the mapping of the abstract model definitions in  CBR cases, similarity function, parameters, variables, and weighting priorities of the deployed CBR system.

The mapping from the abstract model to the concrete model is achieved in the following way:

\begin{itemize}
    \item TD components like attributes, measurements, and tolerances are mapped in the CBR cases description.
    \item General and specific objectives are mapped in similarity function and evaluation function.
    \item Actions are mapped in solutions for the CBR cases.
    \item Premises are mapped in the first cases for the CBR system.
\end{itemize}

The proposed sequence of steps to map from AM to CM model is as follows:

\begin{enumerate}
    \item Mapping of attributes and measurements necessary to achieve the specific objectives;
    \item CBR case description definition using the mapped set of attributes and measurements;
    \item Tolerance level definition for the CBR case components;
    \item Mapping of the actions for the solution of CBR cases;
    \item Similarity function definition;
    \item Evaluation function definition; and
    \item CBR 4R cycle operation process.
\end{enumerate}

The last step, the CBR 4R cycle operation process, is described in this paper as a non-exhaustive sequence of \textit{how to do it} procedures and \textit{hints} that can be used in the CBR 4R cycle.

\subsection{The CBR Case}

\begin{figure}[ht]
\centering
\includegraphics[width=.7\linewidth]{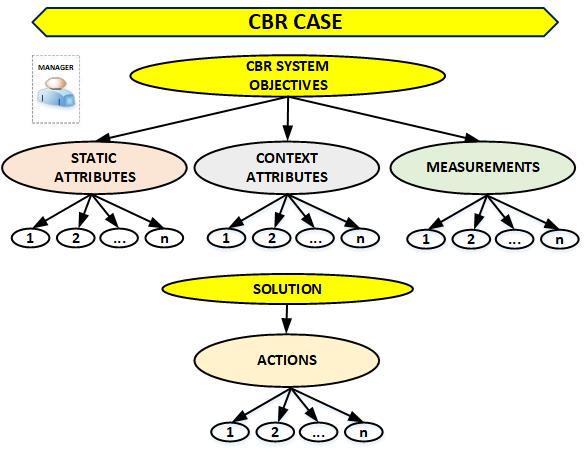}
\caption{A CBR \textit{case}}
\label{fig:case}
\end{figure}

In CBR terminology a \textit{case} is a problem situation composed by a set of parameters describing the problem domain and the associated solution for the problem (Figure \ref{fig:case}) \cite{lamy_explainable_2019}.

A generic CBR case is:
\begin{equation}
C_{j} = (p{_j},a_j)
\end{equation}

where,
\begin{equation}
p_{j} = \{SA{_j}, CA{_j}, M{_j},T{_j}\}
\end{equation}
\begin{equation}
a_{j} = \{a{_{j1}}, a{_{j2}}, ... , a{_{jx}}\}
\end{equation}
\begin{equation}
SA_{j} = \{SA{_{j1}}, SA{_{j2}}, ... , SA{_{jn}}\}
\end{equation}
\begin{equation}
CA_{j} = \{CA{_{j1}}, CA{_{j2}}, ... , CA{_{jk}}\}
\end{equation}
\begin{equation}
M_{j} = \{M{_{j1}}, M{_{j2}}, ... , M{_{jz}}\}
\end{equation}
\begin{equation}
T_{j} = \{T{_{j1}}, T{_{j2}}, ... , T{_{jy}}\}
\end{equation}

The \textit{case} $C_{j} $ is composed by the set of parameters $p_{j}$ and the associated set of actions $a_j$. The set of parameters $p_{j}$ includes all the relevant static attributes ($SA_{j}$), context attributes ($CA_{j}$), measurements ($M_{j}$) and tolerances ($T_{j}$) for the problem situation being described.

The \textit{case} solution is composed by the set of relevant actions $a_j$ to achieve the defined objectives. These actions allow operations or sets of operations that are used to react to a particular problem in the system.

\subsection{Similarity Function (SF)}

Similarity is a crucial aspect of CBR. A similarity function is used to retrieve similar cases from the case-database when a new case or unsolved case arrives at the system. In summary, a new case or unsolved case is the description of a new problem to be solved.

Examples of similarity functions include the identification of similarity based on rules, correlation testing, K-nearest-neighbor (KNN) techniques, and the cosine similarity measure, among others \cite{zhai_associated_2019} \cite{osuszek_case_2015} \cite{watson_case-based_1999} \cite{m._oliveira_evaluating_2017} \cite{lin_case-based_2020}.

The similarity function, regardless of the option, requires the definition of the evaluation indexes and their weights for the given real-world problem. The definition of the similarity measures is an actual challenging research problem and has a significant dependency on the target problem \cite{kwon_preliminary_2019}.

From the methodological point of view, the similarity function is mapped from the specific objectives and is composed of a set of attributes and measurements with weights that define their priority in the definition of the similarity.

The similarity of one case to others can, from the methodological point of view, be defined by averaging the distinct similarities of part of the case, for example, a case with three indexed attributes $ (x {_i}, x {_j}, x {_l}) $ will be similar to another case $ (y {_i}, y {_j}, y {_l}) $ if the attributes $i$, $j$ and $l$ of the cases $ x $ and $ y $ are similar to each other. This partial similarity, by attribute, is called local similarity.

The local similarity is calculated according to the type of attribute that defines the case, and a specific function can be used for each type of data. The functions used for calculate local similarity, as an example, are ladder function, linear function, equality function, maximum function, intersection function, and contrast function \cite{m._oliveira_evaluating_2017}.

The global similarity determines how similar one case is to the other using the values of local similarities to which weights can also be assigned.

Attributes and measurements have direct and indirect relation with the specific objectives. Consequently, two issues arrive in terms of methodologically mapping of the similarity function parameters: i) to consider indirect similarities; and ii) to consider similar cases in distinct contexts. Indirect relations must be indexed, and a similar case in distinct contexts must be differentiated by choosing attribute weights adequately.

\subsection{Evaluation Function (EV)}

The evaluation function is an optional facility that may be included in the CBR operation process. It is not conventionally inserted in the CBR cycle but might be helpful in the CBR operation cycle. The EV is mapped from the objectives, attributes, and tolerances of the concrete model.

The fundamental idea of using an evaluation function is to interpret the state of the measurements, comparing them with the tolerances defined for the CBR system. It provides a kind of on-the-fly evaluation of the CBR system behavior.

This function can be used for two purposes: i) to generate periodic warnings and diagnoses of symptoms and alerts detected in the CBR system; and ii) to check if a solution applied to the CBR system meets the objectives (general and specific) and tolerances defined.

\section{CBR Operation Process - Hints and How to Do It}

After the mapping from the abstract model to the concrete model with the definition of objectives, attributes, measurements, actions, and cases, the CBR 4R cycle starts (Recover, Reuse, Revision and Retain).

For each of these steps, some procedures shall be executed by the CBR system. These procedures depend on the actual problem-solving issue focused on the CBR system, but there are commonalities. We explore some of these commonalities to provide a set of \textit{hints} and possible \textit{how to do actions} for the operation of CBR systems.

\subsection{Recover}

The Cycle 4R has as its starting point the recovery phase that retrieves a similar case and evaluates it. This task can be triggered in two different ways: reactively or proactively.

In reactive mode, an alarm requests analysis triggered by a current problem, to obtain a solution or optimize the system. In proactive mode, the CBR system is activated to check the system's situation and occasionally propose improvements proactively or a solution.

When no case returns from the case-database, two procedures are suggested:
\begin{itemize}
    \item Use the method assisted by the manager, where he provides a new solution.
    \item Use an automated method where the solution is automatically mapped considering the defined objectives or an arbitrary solution is attributed to the current problem. The arbitrary solution attribution corresponds to a brute force learning method.
\end{itemize}

\subsection{Revision}

In the review phase, CBR assesses the efficiency of the proposed solution. For the review, it is necessary to wait a specific time until the actions take effect, and the attributes and measurements of the new state of the system are updated.

In this step, the evaluation function is used to check if the new solution presents improved performance. Without improvements, the new case is considered unsuccessful. The new case, positive or negative, is, by convention, stored in the CBR database.

After applying the solution, the current and previous attributes and measurements, cannot differ much from the previous state. With a considerable variation, it is not possible to identify whether the system has improved due to the solution adopted or simply because the state of the resources has changed. As adopted in other AI techniques, to use a discount is recommended in these cases.

Another recommendation is the creation of a configurable equivalence threshold for the cases. The aim is to avoid many very similar cases populating the case-database and contribute to its excessive growth, which results in a performance problem.

False-positive cases can be negated in the next reuse. For false-negative cases, we suggest a period of validity, both for positive and negative cases, simulating the human forgetfulness of old facts that are rarely used.

\section{Using the Abstract and Concrete Models for an CBR-based Application - Cognitive Management of Bandwidth in Network Links}
\label{sec:CasoUso}

We exemplify now how abstract and concrete models can facilitate the definition of the objectives, attributes, measurements, and tolerances to model a CBR-based application.

The target CBR-based application (BAMCBR) aims to manage bandwidth in links of an MPLS (MultiProtocol Label Switching Network). The link management is executed by a Bandwidth Allocation Model (BAM) that dynamically receives requests for link setup and grants or denies these requests based on the link bandwidth availability in the network \cite{reale_alloctc-sharing:_2011} \cite{martins_uma_2015}. The cognitive management consists of CBR deciding when should the BAM model be changed among a set of options based on network parameters status.  Oliveira in \cite{oliveira_cognitive_2018} has a detailed description of this cognitive management application, and, in this paper, we focus on illustrating how AM and CM can be used to model the CBR application.

It follows the Abstract Model definition and Concrete Model mappings.

\subsection{BAMCBR Abstract Model}

The first step in building the abstract (AM) model is the representation of the technological domain with the definition of its objectives, attributes, measurements, and tolerances.

\subsubsection{BAMCBR Tecnological Domain (TD):}
The CBR system's technological domain is the cognitive management of an MPLS/DS-TE type computer network with bandwidth allocation models (BAMCBR Tool) \cite{oliveira_cognitive_2018}.

\subsubsection{BAMCBR General and Specific Objectives:}
The general objective of the BAMCBR is to decide when should the BAM model be changed among a set of BAM model options available based on network link-state performance parameters and input traffic.

The specific objectives drive the BAM model reconfiguration decision process and are the following:

\begin{enumerate}
    
    \item   To maximize link throughput
    \item   To minimize link preemption; and
    \item   To minimize link devolution.
  \end{enumerate}
  
It is essential to highlight that only with the expertise about the application domain (BAM model operation \cite{martins_uma_2015}) it is possible to know that throughput maximization, preemption, and devolution minimization are management objectives achievable by the reconfiguration of the BAM model \cite{reale_preliminary_2014}.

\subsubsection{BAMCBR Static Atributes:}

The static attributes are the BAM model static configuration parameters. In the BAMCBR they are the total managed  link bandwidth and the configured bandwidth allocate per class of traffic in the link (BC - Bandwidth Constraint \cite{martins_uma_2015}).

\subsubsection{BAMCBR Contextual Atributes:}

The contextual attributes define the management context, and the modification of these attributes, although less frequent, does reflect the manager's perception of what he wants from the CBR system. In BAMCBR, the adopted BAM and the tolerances for throughput, preemption, and devolution are the main contextual parameters. Any change on these attributes implies in restarting the CBR learning process.

\subsubsection{BAMCBR Measurement Attributes:}

The measurement attributes are the link variables that indicate the link's performance and state in a given moment. These are measurable dynamic variables belonging to the target CBR system (network link). Examples of measurement attributes used by the BAMCBR tool are link preemption, link devolution, packet loss, LSP (Label Switched Path) request blocking, and link utilization \cite{oliveira_cognitive_2018}.

\subsubsection{BAMCBR Tolerances:}

The tolerances represent the accepted values range for attributes in general. Their definition requires expertise in the domain area and they refines the specific objectives. In the BAMCBR, as an example, the management accepts a link utilization of 10\% with a 10\% tolerance.

\subsubsection{BAMCBR Actions:}
The actions correspond to the set of operations used to solve a problem for a certain case. In BAMCBR, the action is to reconfigure the current BAM model among the available options: MAM (Maximum Allocation Model), RDM (Russian Dolls Model) and ATCS (AllocTC-Sharing) models \cite{reale_preliminary_2014}.

\subsubsection{BAMCBR Premises:}

The premises are the set of problems or states to which a solution is known. For example, when the current BAM is MAM or RDM, and the utilization is less or equal to 50 \%, the BAM might be reconfigured to ATCS \cite{reale_preliminary_2014}.

\subsection{BAMCBR Concrete Model}

The concrete model (CM) maps the abstract model in the parameters and variables used by the CBR system.

\subsubsection{BAMCBR Case Description:}

\begin{figure*}[ht]
\centering
\includegraphics[width=.7\linewidth]{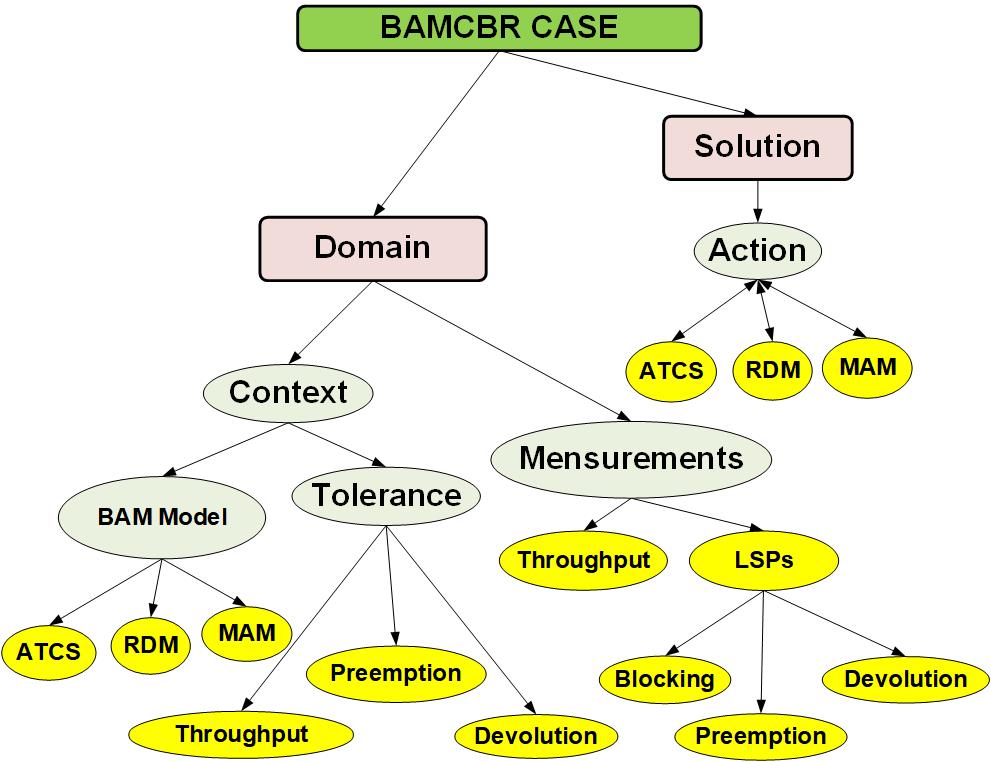}
\caption{Caso BAMCBR}
\label{fig:BAMCBR-CASE}
\end{figure*}

The \textit{case} description includes the attributes, measurements and tolerances previously defined in the abstract model. In BAMCBR, it includes BAM models, measurement variables like preemption, utilization, loss, blocking and devolution, and the action options for BAM model change (Figure \ref{fig:BAMCBR-CASE}).

\subsubsection{BAMCBR Case Solution:}

The actions set for BAMCBR is to dynamically switch between ATCS, RDM and MAM models \cite{reale_preliminary_2014}.

\subsubsection{BAMCBR Evaluation Function:}

The evaluation function verifies the actual state of the managed system by comparing the actual case with previous cases using measurements and tolerances defined. For example, it must also be able to verify if a solution applied to the network is closer to the objective than the solution previously adopted. BAMCBR uses the WkNN function to evaluate the current state of the network in relation to other previous states.

Weights in the evaluation function reflect management expertise. For example, BAMCBR considers that devolution generates a negative impact more significant than preemption, which, in turn, generates a more significant impact than blocking. As such, the weights for devolution, preemption, and blocking are 3, 2, and 1, respectively.

\subsubsection{BAMCBR Similarity Function:}

The functions for local and global similarity defined for the BAMCBR tool are indicated in Table \ref{tab:TrafficProfile2}.

\begin{table}[ht]
\caption{BAMCBR Similarity Functions}
\label{tab:TrafficProfile2}
\begin{center}
\begin{tabular}{|c|c|c|} \hline
Attribute/Measurement & Similarity Function & Weight  \\ \hline
BAM Model & Equal Function & 40 \\
Throughput & Linear Function & 30 \\
Blocking& Linear Function & 30 \\
Devolution& Linear Function & 20 \\
Preemption& Linear Function & 20 \\\hline
\multicolumn{3}{|c|}{Global Similarity Function: WkNN} \\
\multicolumn{3}{|c|}{Cut-off Threshold 96\%} \\
\multicolumn{3}{|c|}{Equivalence Threshold 98.5\%} \\   
\hline
\end{tabular}
\end{center}
\end{table}

The cut-off threshold is used for recovering similar solutions within similarity equal or superior to the indicated limit. The equivalence threshold limit is used to avoid storing multiple nearly identical solutions at the database and, thus, populated it excessively. In other words, solutions with similarity equal or superior to the equivalence threshold limit are not stored in the CBR database.

\section{Final Considerations}\label{sec:conclusion}

The methodological approach to model CBR-based applications uses the definition of an abstract model (AM) that is subsequently mapped in a concrete model (CM). 

The abstract model represents the domain to which CBR is applied and its definition needs essentially the knowledge of an expert in the application domain. The concrete model corresponds to the CBR parameters whose mapping from the abstract model requires CBR expertise. The splitting in two models facilitates the model process and, in addition, allows the allocation of domain and CBR specialists to the distinct phases of the modeling process.

As such, this proposal's inherent advantage is that it allows a task division between specialists in the domain and specialist in CBR. This facilitates the modeling process and has the potential to foster the utilization of CBR in an even large number of areas where computer science expertise is less frequent.


\small
\bibliography{biblio}
\bibliographystyle{unsrt}

\end{document}